\documentclass[10pt, a4paper]{article}
\usepackage{lrec}
\usepackage{graphicx}
\usepackage{tabularx}
\usepackage{soul}
\usepackage{xcolor}
\usepackage{url}
\usepackage{caption}
\usepackage{epstopdf}
\usepackage[latin1]{inputenc}

\usepackage{hyperref}
\usepackage{xstring}

\title{MultiWOZ 2.1: A Consolidated Multi-Domain Dialogue Dataset with State Corrections and State Tracking Baselines}

 \name{Mihail Eric$^{\ast}$, Rahul Goel$^{\ast}$, Shachi Paul \\
 {\bf \large Adarsh Kumar, Abhishek Sethi, Anuj Kumar Goyal, Peter Ku} \\
{\bf \large  Sanchit Agarwal, Shuyang Gao, Dilek Hakkani-T\"ur}}
 \address{ $^{\ast}$ Authors Contributed Equally. \\
   mihaeric@amazon.com, goelrahul@google.com,  shachipaul@google.com \\
   kumar92@wisc.edu, \{abhsethi, anujgoya, kupeter,  agsanchi, shuyag, hakkanit\}@amazon.com}

\abstract{ MultiWOZ 2.0~\cite{Budzianowski2018MultiWOZA} is a recently released
  multi-domain dialogue dataset spanning 7 distinct domains and containing over
  10,000 dialogues. Though immensely useful and one of the largest resources of
  its kind to-date, MultiWOZ 2.0 has a few shortcomings. Firstly, there is
  substantial noise in the dialogue state annotations and dialogue utterances
  which negatively impact the performance of state-tracking models. Secondly,
  follow-up work~\cite{lee2019convlab} has augmented the original dataset with
  user dialogue acts. This leads to multiple co-existent versions of the same
  dataset with minor modifications. In this work we tackle the aforementioned
  issues by introducing MultiWOZ 2.1. To fix the noisy state annotations, we use
  crowdsourced workers to re-annotate state and utterances based on the original
  utterances in the dataset. This correction process results in changes to over
  32\% of state annotations across 40\% of the dialogue turns. In addition, we
  fix 146 dialogue utterances by canonicalizing slot values in the utterances to
  the values in the dataset ontology. To address the second problem, we combined
  the contributions of the follow-up works into MultiWOZ 2.1. Hence, our dataset
  also includes user dialogue acts as well as multiple slot descriptions per
  dialogue state slot. We then benchmark a number of state-of-the-art dialogue
  state tracking models on the MultiWOZ 2.1 dataset and show the joint state
  tracking performance on the corrected state annotations. We are publicly
  releasing MultiWOZ 2.1 to the community, hoping that this dataset resource
  will allow for more effective models across various dialogue subproblems to be
  built in the future. \\ \newline  }

\begin{document}

\maketitleabstract

\section{Introduction}
In task-oriented conversational systems, dialogue state tracking refers to the % RAHUL: why conversational here and not dialogue 
problem of estimating a user's goals and requests at each turn of a
dialogue. The state is typically defined by the underlying ontology of the
domains represented in a dialogue, and a system's job is to learn accurate
distributions for the values of certain domain-specific slots in the
ontology. There have been a number of public datasets and challenges released to
assist in building effective dialogue state tracking modules
~\cite{Williams2013TheDS,Henderson2014TheSD,Wen2017ANE}.

One of the largest resources of its kind is the MultiWOZ 2.0 dataset, which spans 7
distinct task-oriented domains including hotel, taxi, and restaurant booking
among others ~\cite{Budzianowski2018MultiWOZA}. This dataset has been a unique
resource, in terms of its multi-domain interactions as well as slot value
transfers between these domains, and has quickly attracted researchers for
dialogue state tracking~\cite{nouri2018toward,Goel2019HyST,Wu2019TransferableMS} and dialogue policy learning~\cite{ZhaoEskenazi:2019}.

Though the original MultiWOZ 2.0  dataset comes with fine-grained dialogue state
annotations for all the domains at the turn-level, in practice we have found
substantial noise in the annotations of dialogue state values. While some amount
of noise in annotations cannot be avoided, it is desirable to have clean data so
the error patterns in various models can be attributed to model mistakes rather
than the data. 

To this end, we re-annotated states in the MultiWOZ 2.0 dataset with a
different set of interannotators. We specifically accounted for 4 kinds of
common mistakes in MultiWOZ 2.0, detailed in Section \ref{sec:error} In addition,
we also corrected spelling errors and canonicalized entity names as detailed in
Section \ref{user-corr}

Recently there have been a number of extensions to the original MultiWOZ 2.0
 dataset that have added additional annotations such as user dialogue act
 information~\cite{lee2019convlab}. We added this information to our version
 of the dataset so that it has both system and user dialogue acts. Additionally, we added slot-descriptions for all the dialogue state slots present in the dataset, motivated by recent work on low-resource and zero-shot natural language understanding tasks~\cite{Bapna2017TowardsZF,Shah2019RobustZC,RastogiDSTC8}. 

Post-correction, we ran state-of-the-art dialogue state tracking
models on the corrected data to provide competitive baselines for this
new dataset.  With this work, we release the corrected and
consolidated MultiWOZ 2.0 which we call \textit{MultiWOZ 2.1}, as well
as baselines consisting of state-of-the-art dialogue state tracking
techniques on this new data.

In Section \ref{sec:corrections} we provide details for the data correction process
and provide examples and statistics on the corrections. We then detail the slot descriptions we added in Section~\ref{sec:slotdesc} In Section~\ref{sec:da},
we provide the dialogue act statistics for the user dialogue acts included in our dataset. We detail our baseline
models in Section \ref{sec:base} We discuss the performance on this new dataset
in Section \ref{sec:results}

\begin{table}[]
\centering
\begin{tabular}{l|l|l}
{\bf \# Values} & {\bf Previous Value} & {\bf New Value} \\ \hline
6279     & none           & dontcare  \\
2011     & none           & yes       \\
1159     & none           & hotel     \\
1049     & dontcare       & none      \\
920      & none           & centre   
\end{tabular}%
\caption{Top 5 slot value changes (all data) between MultiWOZ 2.1 and MultiWOZ 2.0 by frequency count}
\label{tab:top-changes}
\end{table}

\begin{table}[] 
  \centering
  \begin{tabular}{l|l|l} 

    \textbf{Slot Name}             &  \textbf{2.0} & \textbf{2.1} \\ \hline
    taxi-leaveAt          & 119             & 108                          \\
    taxi-destination      & 277             & 252                          \\
    taxi-departure        & 261             & 254                          \\
    taxi-arriveBy         & 101             & 97                           \\
    restaurant-people     & 9               & 9                            \\
    restaurant-day        & 10              & 10                           \\
    restaurant-time       & 61              & 72                           \\
    restaurant-food       & 104             & 109                          \\
    restaurant-pricerange & 11              & 5                            \\
    restaurant-name       & 183             & 190                          \\
    restaurant-area       & 19              & 7                            \\
    bus-people            & 1               & 1                            \\
    bus-leaveAt           & 2               & 1                            \\
    bus-destination       & 5               & 4                            \\
    bus-day               & 2               & 1                            \\
    bus-arriveBy          & 1               & 1                            \\
    bus-departure         & 2               & 1                            \\
    hospital-department   & 52              & 48                           \\
    hotel-people          & 11              & 8                            \\
    hotel-day             & 11              & 13                           \\
    hotel-stay            & 10              & 10                           \\
    hotel-name            & 89              & 89                           \\
    hotel-area            & 24              & 7                            \\
    hotel-parking         & 8               & 4                            \\
    hotel-pricerange      & 9               & 8                            \\
    hotel-stars           & 13              & 9                            \\
    hotel-internet        & 8               & 4                            \\
    hotel-type            & 18              & 5                            \\
    attraction-type       & 37              & 33                           \\
    attraction-name       & 137             & 164                          \\
    attraction-area       & 16              & 7                            \\
    train-people          & 14              & 12                           \\
    train-leaveAt         & 134             & 203                          \\
    train-destination     & 29              & 27                           \\
    train-day             & 11              & 8                            \\
    train-arriveBy        & 107             & 157                          \\
    train-departure       & 35              & 31                          
  \end{tabular}
  \caption{Comparison of slot value vocabulary sizes (training set) between
    MultiWOZ 2.0 and MultiWOZ 2.1. Note that the vocabulary sizes reduced
    drastically for most slots (except train-arriveby and train-leaveat) due to
    the data cleaning and canonicalization.  }
\label{tab:vocab}
\end{table}

\section{Dataset Corrections} \label{sec:corrections}

\begin{table*}[]
\centering
\resizebox{\textwidth}{!}{%
\begin{tabular}{l|lll}
{\bf Type}              & {\bf Conversation} & {\bf MultiWOZ 2.0} & {\bf MultiWOZ 2.1} \\ \hline
Delayed    & \tt{User: I'd also like to try a Turkish}           &             &             \\
Markups     &  \tt{restaurant. Is that possible?}          &   restaurant.food: None          &    restaurant.food: Turkish         \\
                  &   \tt{Agent: I'm sorry but the only }          &             &             \\
                  &   \tt{restaurants in that part of town serve}          &             &             \\
                  &   \tt{either Asian food or African food.}          &             &             \\ 
                  &   \tt{User: I don't mind changing the area. }            &           &             \\
                  &   \tt{I just need moderate pricing and}            &           &             \\ 
                  &   \tt{want something that serves Turkish food.}          &    restaurant.food: Turkish           &  restaurant.food:Turkish  \\ \hline
Multi            & \tt{User: Can you tell me more about}  &   hotel.name: The Cambridge Belfry          &   hotel.name: The Cambridge Belfry          \\
-annotations     & \tt{Cambridge Belfry}             &   attraction.name: belf          &    attraction.name: None         \\ \hline
Mis              & \tt{User: Yes, I need to leave on}             &          &             \\
-annotations      & \tt{Thursday and am departing}             &  train.leaveAt: Thursday           &    train.leaveAt: None        \\  
                & \tt{from London Liverpool Street.}             &  train.day: Not Mentioned           &     train.day: Thursday        \\  \hline
Typos             &  \tt{Although, I could use some help finding}            &             &             \\
                  &  \tt{an attraction in the centre of town.}             &    attraction.area: cent         &   attraction.area: Centre          \\ \hline
Forgotten       &  \tt{User: No particular price range, but}            &             &            \\
 values          &  \tt{I do need a restaurant that is available  }            &    &            \\ 
                   &  \tt{to book 7 people on Friday at 19:15.}            &  restaurant.pricerange: None           &  restaurant.pricerange: Dontcare          \\ \hline
Value Cano-      &  \tt{User: I think you should try }            &             &            \\
nicalization          &  \tt{again. Cambridge to Bishop  }            &    &            \\ 
                 &  \tt{Stafford on Thursday.}            &  train.destination: Bishop Stortford           &   train.destination: Bishops Stortford          \\ \hline

\end{tabular}%
}
\caption{Examples of annotation errors between MultiWOZ 2.0 and 2.1}
\label{tab:annotation-examples}
\end{table*}

The original MultiWOZ 2.0 dataset was collected using a Wizard-of-OZ setup~\cite{Kelley1984AnID} whereby conversations were conducted between two crowdworkers, one playing the role of the \emph{Wizard} and the other playing the \emph{User}. The \emph{User} was provided with a goal (for e.g. `book a hotel and a taxi to the hotel') and interacted with the \emph{Wizard} with a text-based chat interface to achieve his goal. In the course of a conversation, the \emph{Wizard} had access to a graphical-user-interface connected to a backend database, and they were expected to annotate state information in user utterances using both drop-down menus and free-form text inputs. The use of free-form text inputs made it so that the values annotated by the \emph{Wizard} were not guaranteed to be consistent with the underlying database ontology. This, combined with mistakes made by the crowd-workers resulted in several types of annotation mistakes which we outline below.

\begin{table}[]
  \centering
  \begin{tabular}{l|c}
    %\hline
    {\bf Correction Type }& {\bf \% of Slot Values}\\
    \hline
    no change & 98.16\%\\         %count = 1427708
    \emph{none} $\rightarrow$ value & 1.23\%\\      %count = 31851
    valueA $\rightarrow$ valueB & 0.44\%\\  %count = 611143
    value $\rightarrow$ \emph{none} & 0.17\%\\      %count = 30195
    value $\rightarrow$ \emph{dontcare} & 0.23\%\\      %count = 6279
    %\hline
  \end{tabular}
  \caption{Percentage of values of slots changed in MultiWOZ 2.1 vs. MultiWOZ 2.0}
  \label{tab:corr-type}
\end{table}

\subsection{Dialogue State Error Types} \label{sec:error}
The most common errors types in the original dialogue state annotations include
the following:
\begin{itemize}
\item {\em Delayed markups.} These refer to slot values that were annotated one
  or more turns after the value appeared in the user utterances. Row 1 of
  Table~\ref{tab:annotation-examples} shows this case where the ``Turkish'' value
  appears one turn late in the MultiWOZ 2.0 dialogue.
\item {\em Multi-annotations.} The same value is annotated as belonging to
  multiple slots, usually one of these is correct and the other one is
  spurious. Row 2 of Table~\ref{tab:annotation-examples} shows such a case where
  ``belf'' is spurious.
\item {\em Mis-annotations.} The value is annotated as belonging to a wrong slot
  type. In row 3 of Table~\ref{tab:annotation-examples} we can see a case where
  ``Thursday'' appears in a wrong slot.
\item {\em Typos.} The value is annotated, but it includes a typo or is not
  canonicalized. Row 4 of Table~\ref{tab:annotation-examples} exhibits such a case with
  ``centre'' misspelled.
\item {\em Forgotten values.} The slot value never occurs in the dialogue state,
  even though it was mentioned by the user. Row 5 of
  Table~\ref{tab:annotation-examples} is an example where ``dontcare'' is never
  seen in the data.
\end{itemize}

\subsection{Dialogue State Corrections}
Our corrections were of two types: manual corrections and automated
corrections. Manual corrections involved asking annotators to go over each
dialogue turn-by-turn and correcting mistakes detected in the original
annotations. During this step, we noticed that sometimes the dialogue state
could include multiple values, and hence we annotated them as
such. Table~\ref{tab:multi-value-examples} includes examples of these cases.
MultiWOZ 2.1 has over 250 such multi-value slot values.

After the first manual pass of annotation correction, we wrote scripts to
canonicalize slot values for lookup in the domain databases provided as part of
the corpus. Row 6 of Table~\ref{tab:annotation-examples} shows one such example.
We also present some of the most frequent corrections for state values in Table
\ref{tab:top-changes}. Table~\ref{tab:corr-type} presents statistics on the types
of corrections made.

Due to our canonicalization and reannotation, the vocabulary sizes of many of the
slots decreased significantly (Table~\ref{tab:vocab}) except 2 slots -
``train-leaveAt'' and ``train-arriveBy".  For these slots we noticed that there
were times missing in the dialogue states (such as ``20:07'') which our annotations additionally
introduced. We also canonicalized all times in the 24:00 format.

\begin{table}
  \begin{small}
    \begin{tabular}{|l|}
      \hline 
          {\tt Agent: I have two restaurants. They }\\
          {\tt        are Pizza Hut Cherry Hinton and }\\
          {\tt        Restaurant Alimentum.}\\
          {\tt User:  What type of food do each }\\
          {\tt        of them serve?}\\[0.3em]
          {\bf restaurant.name}: \emph{Pizza Hut Cherry Hinton},\\
          \emph{Restaurant Alimentum}\\
          \hline
              {\tt User: I would like to visit a museum }\\
              {\tt       or a nice nightclub in the north.}\\[0.3em]
              {\bf attraction.type}: \emph{museum, nightclub}\\
              \hline
                  {\tt User: I would also like a reservation  }\\
                  {\tt       at a Jamaican restaurant in that area }\\
                  {\tt       for seven people at 12:45, if there}\\
                  {\tt       is none Chinese would also be good.}\\[0.3em]
                  {\bf restaurant.food}: \emph{Jamaican (preferred), Chinese}\\[0.3em]
                  \hline
                      {\tt User: I would prefer one in the cheap}\\
                      {\tt       range, a moderately priced one is }\\ 
                      {\tt       fine if a cheap one isn't there.}\\[0.3em]
                      {\bf restaurant.pricerange}: \emph{cheap (preferred), moderate}\\
                      \hline
    \end{tabular}
  \end{small}
  \caption{Example dialogue sections with multi-value slots in their states.}
  \label{tab:multi-value-examples}
\end{table}

\subsection{Dialogue Utterance Corrections} \label{user-corr}
It is often the case when building dialogue state systems that the target slot
values are mentioned verbatim in the dialogue history. Many copy-based dialogue
state tracking models heavily rely on this
assumption~\cite{goel2018flexible}. In these situations, it is crucial that the
slot values are represented correctly within the user and system
utterances. However, because dialogue datasets are often collected via
crowdsourced platforms where workers are asked to provide utterances via
free-form text inputs, these slot values within the utterances may be misspelled
or they may not be consistent with the true values from the ontology.

To detect potential error cases within the utterances, for every single
dialogue turn, we computed the terms that have Levenshtein distance less than 3
from the slot values annotated for that turn. We then performed string matching
for these terms within the turn, forming a set of \emph{error candidates}. This
created a candidate set of 225 potential errors which we then manually inspected
to filter out those candidates which were false positives, leaving a collection
of 67 verified errors. We then programmatically scanned the entire dataset
applying corrections to the verified errors, changing 146 total utterances. 

As an example of a corrected utterance: \emph{``I'm leaving from camgridge and
  county folk museum.''} was changed to \emph{``I'm leaving from cambridge and
  county folk museum.''} Without such a correction, it would be very difficult
for a span-based copy mechanism to correctly identify the slot value "cambridge and county folk museum" in its original form.

\section{Slot Description} \label{sec:slotdesc}

Recent works in low-resource cross-domain natural language understanding~\cite{Bapna2017TowardsZF,Shah2019RobustZC,RastogiDSTC8} have developed alternative techniques for building domain-specific modules without the need for many labeled or unlabeled examples. In the case of slot-filling and dialogue state tracking systems, these works have shown that new domains can be bootstrapped using only slot descriptions via learned latent semantic representations. These are very promising techniques as they allow systems to scale to new schemas and ontologies without extensive data annotation. 

To help encourage further research in these techniques, we had two annotators each add at least one natural language description for each slot in MultiWOZ 2.0. Models that use these more detailed descriptions of slot semantics may be able to achieve increased accuracy, especially in domains with little or no data and cases where the slot names alone aren't very meaningful or precise. This setting may be representative of real-world applications, and this data enables experimentation with zero or few-shot methods. Examples of our slot descriptions are presented in Table ~\ref{tab:slotdesceg}.

\begin{table}[] 
  \centering
  \begin{tabular}{l|l} 
\textbf{System Dialogue Act} &  \textbf{Frequency} \\ \hline
Train-OfferBook & 3032 \\
Restaurant-Inform & 8066 \\
Hotel-Request & 3213 \\
general-reqmore & 13769 \\
Booking-Book  & 5253 \\
Restaurant-NoOffer &  1452 \\
Hotel-NoOffer & 914 \\
Hotel-Inform  & 8222 \\
Booking-NoBook  & 1313 \\
Restaurant-Request  & 3079 \\
Hotel-Select  & 1005 \\
Restaurant-Recommend  & 1495 \\
Attraction-NoOffer  & 490 \\
Hotel-Recommend & 1501 \\
Hospital-Request  & 78 \\
Restaurant-Select & 917 \\
Attraction-Select & 438 \\
Booking-Request &2708 \\
Train-Inform  &7203 \\
Train-OfferBooked &2308 \\
general-bye &9105 \\
Taxi-Request  &1613 \\
Attraction-Recommend  &1451 \\
Train-Request &5520 \\
general-greet &2021 \\
general-welcome &4785 \\
Taxi-Inform &2087 \\
Booking-Inform  &5701 \\
Attraction-Request  &1640 \\
Attraction-Inform &6973 \\
Train-NoOffer &117 \\
Police-Inform &434 \\
Hospital-Inform &515 \\
Train-Select  &389 \\
\end{tabular}
\caption{System dialogue act statistics.  }
\label{tab:dastatssys}
\end{table}

\begin{table}[] 
  \centering
  \resizebox{\columnwidth}{!}{
  \begin{tabular}{|l|l|} 
  \hline
\textbf{Slot} &  \textbf{Description} \\ \hline
attraction-type & \emph{type of the attraction place};\\
& \emph{type of attraction or point of interest} \\
hotel-name &\emph{name of the hotel}; \\
& \emph{what is the name of the hotel}\\\hline
\end{tabular}}
\caption{Example of slot descriptions. These were collected manually for all the slots present in MultiWOZ 2.1 }
\label{tab:slotdesceg}
\end{table}

\begin{table}[] 
  \centering
  \begin{tabular}{l|l} 
\textbf{User Dialogue Act} &  \textbf{Frequency} \\ \hline
Attraction-Inform & 5025 \\
Restaurant-Request & 2750 \\
general-bye & 1097 \\
general-thank & 10493 \\
Restaurant-Inform & 12784 \\
Hotel-Request & 2229 \\
Police-Request  & 179 \\
Police-Inform & 175 \\
Hospital-Request  & 259 \\
Train-Request & 2588 \\
general-greet & 120 \\
Taxi-Inform & 3269 \\
Taxi-Request  & 426 \\
Hotel-Inform  & 12876 \\
Hospital-Inform & 330 \\
Train-Inform  & 11154 \\
Attraction-Request  & 3709 \\
\end{tabular}
\caption{User dialogue act statistics. These were generated automatically using heuristics. }
\label{tab:dastatsuser}
\end{table}

\section{Dialogue act annotation} \label{sec:da}

MultiWOZ 2.0 has annotations for the system dialogue acts but lacks annotation for the user utterances in the dialogue. \cite{lee2019convlab} released a version of the dataset with additional annotations for user dialogue acts. They performed this annotation automatically using heuristics that track the dialogue state, user goal, user utterance, system response and system dialogue act. We used their annotation pipeline to annotate our dataset with dialogue acts. Table ~\ref{tab:dastatsuser} and ~\ref{tab:dastatssys} list the statistics of user and system dialogue acts present in the dataset, respectively.

\section{Baseline Models} \label{sec:base}
Within dialogue state tracking, there are two primary classes of models:
$fixed\;vocabulary$ and $open\;vocabulary$. In $fixed\;vocabulary$ models, the
state tracking mechanism operates on a predefined ontology of possible slot
values, usually defined to be the values seen in the training and validation
data splits. These models benefit from being able to fluidly predict values that
aren't present in a given dialogue history but suffer from the rigidity of
having to define the potentially large slot value list per domain during the
model training phase. By contrast $open\;vocabulary$ models are able to flexibly
extract slot values from a dialogue history but struggle to predict slot values
that have not been seen in the history.

In order to benchmark performance on our updated dataset, we provide joint
dialogue state accuracies for a number of $fixed\;vocabulary$ and
$open\;vocabulary$ models which are reported in Table 4. For the models, the
dialogue history up to turn $n$ is defined as $(u_1, s_1, u_2, s_2, ...,
u_{n-1}, s_{n-1}, u_n)$, where $u_i$ and $s_i$ are the user and system
utterances at turn $i$ respectively. Note that this history also includes the
$n^{\textrm{th}}$ user utterance.

The \emph{Flat Joint State Tracker} refers to a bidirectional LSTM network that
encodes the full dialogue history and then applies a separate feedforward
network to the encoded hidden state for every single state slot. In practice
this amounts to 37 separately branching feedforward networks that are trained
jointly. The \emph{Hierarchical Joint State Tracker} incorporates a similar
architecture but instead encodes the history using a hierarchical network in the
vein of ~\cite{Serban2016BuildingED}. \emph{TRADE} is a recently proposed model
that achieved state-of-the-art results on the original MultiWOZ 2.0 data, using
a generative state tracker with a copy mechanism
~\cite{Wu2019TransferableMS}. The \emph{DST Reader} is a newly proposed model
that frames state tracking as a reading comprehension problem, learning to
extract slot values as spans from the dialogue history ~\cite{Natasha2019}. The
\emph{HyST} is another new model which combines a hierarchical encoder
$fixed\;vocabulary$ system with an $open\;vocabulary$ n-gram copy-based system
~\cite{Goel2019HyST}.

\begin{table} 
  \begin{tabular}{l|c|c}
    %\hline
    \textbf{Model}  & \textbf{MultiWOZ 2.0} & \textbf{MultiWOZ 2.1} \\
    \hline
    FJST          & 40.2\%             & 38.0\%                          \\
    %\hline
    HJST & 38.4\%             & 35.55\%                         \\
    %\hline
    TRADE & \bf{48.6\%}  & 
    \bf{45.6\%}                         \\
    %\hline
    DST Reader  & 39.41\% & 36.4\%                          \\
    %\hline
    HyST & 42.33\% & 38.1\%                          \\
  \end{tabular}
  \caption{Test set joint state accuracies for various models on the MultiWOZ
    2.0 and MultiWOZ 2.1 data. FJST refers to the Flat Joint State Tracker, and
    HJST refers to the Hierarchical Joint State Tracker.}
  \label{tab:results}
\end{table}

\section{Results and Discussion} \label{sec:results}
As we can see from Table~\ref{tab:results}, the relative performances of the
models have remained the same across the data updates. However, we also noticed a consistent drop in
performance for all models on MultiWOZ 2.1 compared to MultiWOZ 2.0, which was a particularly surprising result.

In order
to understand the source of this drop, we investigated the performances of the
Flat Joint State Tracker and Hierarchical Joint State Tracker on the MultiWOZ 2.0 and the MultiWOZ 2.1
datasets. Across the two datasets, we observed that there are 937 new turn-level
prediction errors that the Flat Joint State Tracker makes on MultiWOZ 2.1 that it did not make on
MultiWOZ 2.0. This constitutes 1370 total slot value prediction errors across
the turns. Of these slot value errors, we saw that 184 errors (${\sim}13.4\%$) are
a result of a \emph{dontcare} target label for which our model predicts another
value. 

When we looked at predictions of the Hierarchical Joint State Tracker, we saw that a model trained on MultiWOZ 2.0 generated 331 errors for which the ground truth label was \emph{dontcare} but it predicted \emph{none}. Meanwhile a model trained on MultiWOZ 2.1 generated 748 such errors, a factor increase of over 2.25x. 
As shown in Table~\ref{tab:corr-type}, ${\sim}11.1\%$ of our corrections
involved changing a value to a \emph{dontcare} label so we hypothesize that our
corrections have increased the complexity of learning the \emph{dontcare} label correctly. Given that building systems that can effectively capture user ambiguity is an important characteristic of conversational systems, this leaves ample room for improvement in future models. 

Also noteworthy is the fact that 439 new errors for the Flat Joint State Tracker (${\sim}32.0\%$) are caused when the target
label is \emph{none} but the model predicts another value. As Table~\ref{tab:corr-type} shows
${\sim}8.2\%$ of our corrections involved changing a slot from a value to
\emph{none}, suggesting that MultiWOZ 2.1 now more heavily penalizes spurious slot value
predictions. 

For the Flat Joint State Tracker, we also observed that the largest slot accuracy decrease from MultiWOZ 2.0 to MultiWOZ 2.1 occurred for the {\bf restaurant.name} slot ($87.02\% \rightarrow 83.33\%$). We inspected the kinds of errors the model was generating and found that the vast majority of these errors were legitimate model prediction mistakes on correctly annotated dialogue states. This encourages further research in enhancing the performance of these state-tracking models, especially on proper name extraction.

\section{Conclusion}
We publicly release state corrected MultiWOZ 2.1 and rerun competitive state
tracking baselines on this dataset. The dataset will be available on the MultiWOZ Github repository\footnote{https://github.com/budzianowski/multiwoz/tree/master/data}.  We hope that the cleaner data allows for better model
and performance comparisons on the task of multi-domain dialogue state tracking as well as other dialogue subproblems. \\ \\

% \nocite{*} 
\section{Bibliographical References}
\label{main:ref}

\bibliographystyle{lrec}
\bibliography{lrec2020W-xample}

%% \captionsetup{type=table}
%% \captionsetup{name=\textbf{Appendix}}

\begin{table*}[]
  \centering
  \resizebox{\textwidth}{!}{%
    \begin{tabular}{l|cccccc}
      \textbf{Slot Names}            & \textbf{\% changed } & \textbf{ \# changed } & \textbf{ \% changed } & \textbf{\# changed } & \textbf{\% changed } & \textbf{\# changed } \\ 
                  & \textbf{ Train} & \textbf{ Train} & \textbf{  Dev} & \textbf{ Dev} & \textbf{ Test} & \textbf{ Test} \\ \hline
      taxi-leaveAt          & 0.43\%           & 246              & 0.30\%         & 22             & 0.73\%          & 54              \\
      taxi-destination      & 1.46\%           & 830              & 1.33\%         & 98             & 1.38\%          & 102             \\
      taxi-departure        & 1.47\%           & 833              & 1.29\%         & 95             & 1.41\%          & 104             \\
      taxi-arriveBy         & 0.29\%           & 167              & 0.26\%         & 19             & 0.43\%          & 32              \\
      restaurant-people     & 0.74\%           & 423              & 0.64\%         & 47             & 0.71\%          & 52              \\
      restaurant-day        & 0.72\%           & 410              & 0.62\%         & 46             & 0.68\%          & 50              \\
      restaurant-time       & 0.74\%           & 422              & 0.71\%         & 52             & 0.77\%          & 57              \\
      restaurant-food       & 2.77\%           & 1574             & 2.45\%         & 181            & 2.13\%          & 157             \\
      restaurant-pricerange & 2.36\%           & 1338             & 1.83\%         & 135            & 2.71\%          & 200             \\
      restaurant-name       & 8.20\%           & 4656             & 5.84\%         & 431            & 9.58\%          & 706             \\
      restaurant-area       & 2.34\%           & 1328             & 1.55\%         & 114            & 2.75\%          & 203             \\
      bus-people            & 0.00\%           & 0                & 0.00\%         & 0              & 0\%             & 0               \\
      bus-leaveAt           & 0.00\%           & 0                & 0.00\%         & 0              & 0\%             & 0               \\
      bus-destination       & 0.00\%           & 0                & 0.00\%         & 0              & 0\%             & 0               \\
      bus-day               & 0.00\%           & 0                & 0.00\%         & 0              & 0\%             & 0               \\
      bus-arriveBy          & 0.00\%           & 0                & 0.00\%         & 0              & 0\%             & 0               \\
      bus-departure         & 0.00\%           & 0                & 0.00\%         & 0              & 0\%             & 0               \\
      hospital-department   & 0.12\%           & 68               & 0.00\%         & 0              & 0\%             & 0               \\
      hotel-people          & 1.06\%           & 603              & 0.61\%         & 45             & 0.61\%          & 45              \\
      hotel-day             & 1.00\%           & 565              & 0.69\%         & 51             & 0.65\%          & 48              \\
      hotel-stay            & 1.18\%           & 671              & 0.61\%         & 45             & 0.84\%          & 62              \\
      hotel-name            & 6.90\%           & 3917             & 5.84\%         & 431            & 5.81\%          & 428             \\
      hotel-area            & 3.43\%           & 1947             & 2.03\%         & 150            & 3.95\%          & 291             \\
      hotel-parking         & 2.69\%           & 1526             & 2.78\%         & 205            & 2.67\%          & 197             \\
      hotel-pricerange      & 3.09\%           & 1753             & 2.18\%         & 161            & 2.39\%          & 176             \\
      hotel-stars           & 1.69\%           & 962              & 1.38\%         & 102            & 1.95\%          & 144             \\
      hotel-internet        & 2.27\%           & 1290             & 2.17\%         & 160            & 3.05\%          & 225             \\
      hotel-type            & 3.58\%           & 2035             & 2.64\%         & 195            & 2.79\%          & 206             \\
      attraction-type       & 4.57\%           & 2594             & 4.43\%         & 327            & 4.03\%          & 297             \\
      attraction-name       & 5.99\%           & 3400             & 6.60\%         & 487            & 8.86\%          & 653             \\
      attraction-area       & 2.13\%           & 1212             & 1.79\%         & 132            & 3.23\%          & 238             \\
      train-people          & 0.92\%           & 520              & 0.53\%         & 39             & 0.75\%          & 55              \\
      train-leaveAt         & 2.07\%           & 1178             & 2.12\%         & 156            & 4.64\%          & 342             \\
      train-destination     & 0.91\%           & 518              & 0.69\%         & 51             & 0.87\%          & 64              \\
      train-day             & 0.84\%           & 476              & 0.54\%         & 40             & 0.85\%          & 63              \\
      train-arriveBy        & 1.29\%           & 730              & 1.06\%         & 78             & 2.82\%          & 208             \\
      train-departure       & 1.01\%           & 573              & 0.94\%         & 69             & 0.66\%          & 49              \\ \hline
      Joint                 & 41.34\%          & 23473            & 37.96\%        & 2799           & 45.02\%          & 3319           
    \end{tabular}%
  }
  \caption*{\textbf{Appendix A}: Percentage of changes in dialogue state values before and after
    annotations. The highest number of changed values are in name slots
    (e.g., \emph{restaurant-name, attraction-name, and hotel-name}). Such slots had
    particularly large numbers of spelling mistakes (e.g., \emph{shanghi family
      restaurant} to \emph{shanghai family restaurant}). Note that while the
    number of changes to individual slots is small, we ended up changing the
    joint dialogue state for over 40\% of dialogue turns.} \label{change}
\end{table*}
\end{document}